\documentclass[letterpaper, 10 pt, conference]{ieeeconf}  

\usepackage{graphicx}
\usepackage{float}
\usepackage{amsmath}
\usepackage{threeparttable}

\IEEEoverridecommandlockouts                              

\floatstyle{ruled}
\newfloat{algorithm}{tbp}{loa}
\providecommand{\algorithmname}{Algorithm}
\floatname{algorithm}{\protect\algorithmname}

\usepackage{subfigure}

\let\vec\boldvec

\usepackage{algpseudocode}
\usepackage{epstopdf}
\usepackage{multicol}
\usepackage{booktabs}

\usepackage[hidelinks]{hyperref}
\usepackage{amssymb}

\title{\LARGE \bf
	Generalized Task-Parameterized Skill Learning
}

\author{Yanlong Huang, Jo\~{a}o Silv\'{e}rio, Leonel Rozo, and Darwin G. Caldwell 
	\thanks{All authors are with Department of Advanced Robotics, Istituto Italiano di Tecnologia,
		Via Morego 30, 16163 Genoa, Italy, {\tt\small firstname.lastname@iit.it}}
\thanks{This work was supported by the Italian Ministry of Defense.}}

\begin{document}

\maketitle
\thispagestyle{empty}
\pagestyle{empty}
	
\begin{abstract}
Programming by demonstration has recently gained much attention due to its user-friendly and natural way to transfer human skills to robots. In order to facilitate the learning of multiple demonstrations and meanwhile generalize to new situations, a task-parameterized Gaussian mixture model (TP-GMM) has been recently developed. This model has achieved reliable performance in areas such as human-robot collaboration and dual-arm manipulation. However, 
the crucial task frames and associated parameters in this learning framework are often set by the human teacher, which renders 
three problems that have not been addressed yet: \emph{(i)} task frames are treated equally, without considering their individual importance, \emph{(ii)} task parameters are defined without taking into account additional task constraints, such as robot joint limits and motion smoothness, and \emph{(iii)} a fixed number of task frames are pre-defined regardless of whether some of them may be redundant or even irrelevant for the task at hand.   
In this paper, we generalize the task-parameterized learning by addressing the aforementioned problems. Moreover, we provide a novel learning perspective which allows the robot to refine and adapt previously learned skills in a low dimensional space. Several examples are studied in both simulated and real robotic systems, showing the applicability of our approach.	
		
\end{abstract}

\section{Introduction}	
\label{sec:intro}
As an intuitive and user-friendly way to endow a robot with
skills from humans, Programming by Demonstration (PbD) has become appealing in the past few years \cite{Argall}.
The basic idea of PbD is to extract the important or consistent  
features from demonstrations and then adapt them to 
various situations, which is also referred to as generalization. 
In practice, a myriad of robot tasks are formulated as a regression problem, 
e.g., a mapping from sensory information to robot (motor) actions. 
However, typical regression approaches such as locally weighted regression (LWR) \cite{Atkeson} or Gaussian process regression (GPR) \cite{Rasmussen} may suffer from limited extrapolation capabilities \cite{Calinon2014}.
In order to adapt learned robot skills to 
a broader range of task instances, 
a multi-frame based probabilistic learning framework TP-GMM
was proposed \cite{Calinon2014}. This approach
exploits locally consistent features among demonstrations in different local coordinate systems 
instead of using a single global reference frame, and subsequently transfers local features to new task frames (which describe new task situations), yielding reliable performance for both interpolation and extrapolation.

However,  
the crucial task frames and associated parameters in TP-GMM are usually set according to the human knowledge about the task, which renders three main limitations: \emph{(i)} task frames are treated equally without considering their individual importance. However, depending on human interpretation of tasks, 
the task frames influence may vary over time,
which can be interpreted as the expertise or \emph{confidence} that a specific frame has with respect to a portion of the task, which is overlooked in \cite{Calinon2014};
\emph{(ii)} task parameters are defined regardless of additional task constraints, such as robot joint limits and motion smoothness. These new constraints demand the robot to adapt the learned task-parameterized skill according to additional requirements while performing successfully; \emph{(iii)} a fixed number of task frames are pre-defined ignoring whether some of them are redundant or even irrelevant for the task at hand. These unnecessary frames will increase the computational burden and potentially degrade the overall performance of TP-GMM. 

Besides the foregoing problems, it is worth mentioning that 
human demonstrations might not be optimal for the robot. Namely, the demonstrator may mainly focus on the  
task at hand while the robot capability is not fully exploited, which may lead to high energy movements, unnecessary large joint displacements, or high torque motion, among other problems. 
Also, due to the complicated structure or non-linearity exhibited in the demonstrated trajectories, it is non-trivial to optimize these trajectories effectively. We here propose to take advantage of 
the task-parametrized formulation of TP-GMM by optimizing task parameters instead of directly modifying the model parameters (i.e., GMM means and covariance matrices), while the latter is conventionally done \cite{Guenter}. A clear advantage of our approach is that task parameters lie in a lower dimensional space compared to that of trajectory model parameters.

\begin{figure*}[bt] \centering 
	\includegraphics[width=12.5cm,bb=0cm 3.4cm 28cm 18.6cm,clip]{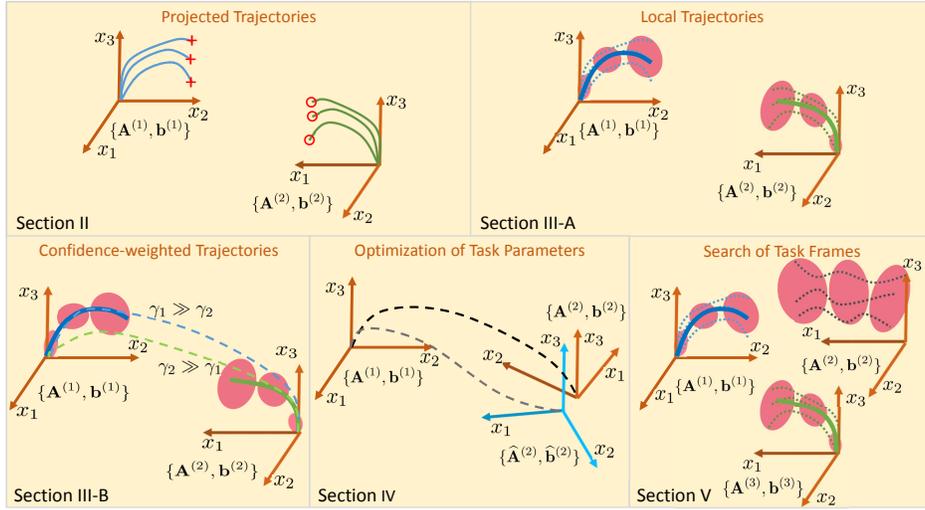}
	\caption{Illustrations of task-parameterized movement learning. \emph{Top-left} plot depicts projected trajectories in different task frames, where `o' and `+' denote the start and end points of trajectories, respectively. \emph{Top-right} plot shows trajectory encoding using GMM and trajectory retrieval using GMR, where the ellipses depict GMM components and the solid curves represent the mean trajectories retrieved by GMR. \emph{Bottom-left} graph illustrates the trajectory generation using the confidence-weighted scheme.  \emph{Bottom-middle} graph presents the trajectory adaptation using the optimized task parameters. \emph{Bottom-right} figure shows a case where the selection of task frames is important.}
	\label{fig:gmm:gmr}
\end{figure*}

In this paper, we first briefly introduce TP-GMM (Section~\ref{sec:tp_gmm:overview}). Subsequently, we consider a variant of TP-GMM (Section~\ref{subsec:tmp}) so as to address the stated problems properly. Using this new formulation, we 
propose a confidence-weighted scheme to address problem \emph{(i)} (Section~\ref{subsec:cw:tmp}). In order to cope with problem \emph{(ii)}, we formulate the optimization of task parameters as a reinforcement learning (RL) problem (Section~\ref{subsec:rl_tp}), with the aim of enabling the robot to finish the task while satisfying additional task constraints. Also, we provide a dual perspective to show that the optimization of task parameters in a lower dimensional space is equivalent to that of model parameters in a higher dimensional space (Section~\ref{subsec:dual}). Furthermore, as a solution to problem \emph{(iii)}
we propose an iterative frame selection algorithm to exploit the most relevant task frames (Section~\ref{sec:forward:search}), where the task parameters optimization is used. Finally, we evaluate our approaches through several examples in Section~\ref{sec:evaluations} and conclude this paper in Section~\ref{sec:conclusion}. An overview of our main contributions is illustrated in Fig.~\ref{fig:gmm:gmr}.

\section{An Overview of Task-parameterized Gaussian Mixture Model \label{sec:tp_gmm:overview}} 
In the context of imitation learning, one crucial ingredient is the consistent features underlying human demonstrations \cite{Calinon2014, Guenter, Ijspeert}. In order to facilitate the extraction of consistent features, 
TP-GMM has been exploited in numerous applications, e.g., human-robot collaborative transportation \cite{Leonel} as well as robot bimanual sweeping \cite{Joao}. Often, a set of candidate task frames (e.g., frames at target objects \cite{Leonel} or robot end-effectors \cite{Joao}) needs to be pre-defined for the implementation of TP-GMM.

Formally, let us consider $P$ \emph{task frames},
and refer to the rotation matrix  $\vec{A}_{t}^{(j)}$ and translation vector $\vec{b}_{t}^{(j)}$ of each frame $\{j\}$ with respect to the global reference frame $\{O\}$ as the \emph{task parameters}, where $t$ denotes the time step and $j=1,2,\ldots,P$. 
We then project human demonstrations $\{\{\vec{\xi}_{t, m}\}_{t=1}^{N}\}_{m=1}^{M}$ into each frame separately and subsequently exploit the local features in different frames. Here, $N$ and $M$ respectively represent the time length of each demonstration and the number of demonstrations, while $\vec{\xi}_{t,m} \in \mathbb{R}^{D}$ represents a $D$-dimensional trajectory point.
The projected trajectories in each frame $\{j\}$ are computed by (see \cite{Calinon2014} for details)
\begin{equation}
\vec{\xi}_{t,m}^{(j)}= (\vec{A}_{t}^{(j)})^{-1}   (\vec{\xi}_{t,m}-\vec{b}_{t}^{(j)}). 
\label{equ:project}
\end{equation}    
If we consider the estimation of consistent features among the projected trajectories from a probabilistic perspective, GMM can be employed \cite{Calinon2014, Leonel, Joao, Muhlig}, which has shown reliable modeling of joint distribution of trajectories.
By using \emph{Expectation Maximization} (EM) algorithm, GMM parameters $\{
\pi_k, \{\vec{\mu}_{k}^{(j)}, \vec{\Sigma}_{k}^{(j)} \}_{j=1}^{P} 
\}_{k=1}^{K}$ in different frames can be estimated, where $K$ represents the number of Gaussian components, $\pi_k$, $\vec{\mu}_{k}^{(j)}$ and $\vec{\Sigma}_{k}^{(j)}$ respectively denote mixture coefficients, Gaussian centers and covariance matrices in each frame $\{j\}$.

By using affine transformations and product of Gaussians,
new GMM components $\{ \pi_k, {\vec{\mu}}_{k,t}, {\vec{\Sigma}}_{k,t} \}_{k=1}^{K}$ at time $t$ in the global frame $\{O\}$ can be computed as
\begin{equation}
\mathcal{N} \!({\vec{\mu}}_{k,t},{\vec{\Sigma}}_{k,t} )\! \propto \! \prod_{j=1}^{P}
\mathcal{N} \! \left( {\vec{A}}_{t}^{(j)} \vec{\mu}_{k}^{(j)}\! + \!{\vec{b}}_{t}^{(j)}, {\vec{A}}_{t}^{(j)} \vec{\Sigma}_{k}^{(j)} ({\vec{A}}_{t}^{(j)})^{T} \right),
\label{equ:reproduce}
\end{equation}
which yields a distribution $\vec{\xi}_t \sim \sum_{k=1}^{K} \pi_k \mathcal{N}({\vec{\mu}}_{k,t}, {\vec{\Sigma}}_{k,t})$ in the frame $\{O\}$. 
Furthermore,
we can decompose $\vec{\xi}$ into input ${\vec{\xi}}_{\mathcal{I}}$ and output $\vec{\xi}_{\mathcal{O}}$, and subsequently, generate a trajectory in the global frame $\{O\}$ as
$\vec{\xi}_{t,\mathcal{O}} | \vec{\xi}_{t,\mathcal{I}} \sim \mathcal{N}( \vec{\mu}_{t,\mathcal{O}}, \vec{\Sigma}_{t,\mathcal{O}})$ by using Gaussian mixture regression (GMR) \cite{Calinon2014, Cohn}.
To name an example, 
if we consider $\vec{\xi}_{\mathcal{I}}$ and $\vec{\xi}_{\mathcal{O}}$ as time and a 3-D trajectory point, respectively, then a sequence of trajectory points in frame $\{O\}$ at different time steps can be generated.
Note that the input is not limited to be time, other inputs can be possible depending on the task characteristics.

\section{Confidence-weighted Task-parameterized Movement Learning \label{sec:tmp}} 
In order to formulate the confidence assignments to task frames, optimization of task parameters as well as frame selection, we first introduce a variant of TP-GMM in Section~\ref{subsec:tmp}. Note that intuitive insights on this variant have been studied for two frames \cite{Calinon2016} and multiple frames \cite{Alizadeh}, while here we aim to provide a mathematical proof. 
Subsequently, we propose a novel confidence-weighted scheme
(described in Section~\ref{subsec:cw:tmp}) which, for example, allows the demonstrator to include information about his/her confidence regarding the relevance/influence of each task frame with respect to the task that is being learned. 
\subsection{Task-parameterized Movement Trajectories \label{subsec:tmp}}
Assuming that we can access to the local GMM models in different task frames, the local trajectory distribution in each frame $\{j\}$ can be represented as $\vec{\xi}^{(j)}\sim\sum_{k=1}^{K} \pi_k \mathcal{N}(\vec{\mu}_{k}^{(j)}, \vec{\Sigma}_{k}^{(j)})$. 
By decomposing
$\vec{\xi}^{(j)}$ as the input ${\vec{\xi}}_{\mathcal{I}}^{(j)}$ and the output $\vec{\xi}_{\mathcal{O}}^{(j)}$, we can generate a local trajectory in frame $\{j\}$ as
$\vec{\xi}_{t,\mathcal{O}}^{(j)} | \vec{\xi}_{t,\mathcal{I}}^{(j)} \sim~ \mathcal{N}( \vec{\mu}_{t,\mathcal{O}}^{(j)}, \vec{\Sigma}_{t,\mathcal{O}}^{(j)} )$ using GMR.
The global trajectory can be viewed as a trade-off among all local trajectories. Formally, the global trajectory $\vec{\xi}_{t,\mathcal{O}}$ can be estimated by maximizing  
$
\prod_{j=1}^{P} \mathcal{P} \left( \vec{\xi}_{t,\mathcal{O}} | \vec{A}_{t,\mathcal{O}}^{(j)} \vec{\mu}_{t,\mathcal{O}}^{(j)}+\vec{b}_{t,\mathcal{O}}^{(j)}, \vec{A}_{t,\mathcal{O}}^{(j)} \vec{\Sigma}_{t,\mathcal{O}}^{(j)} (\vec{A}_{t,\mathcal{O}}^{(j)})^{T} \right)$.
Through the logarithmic transformation, 
this objective can be solved by minimizing the cost function 
\begin{equation}
\begin{aligned}
J(\vec{\xi}_{t,\mathcal{O}})&=\sum_{j=1}^{P} (\vec{\xi}_{t,\mathcal{O}}- \vec{A}_{t,\mathcal{O}}^{(j)} \vec{\mu}_{t,\mathcal{O}}^{(j)}-\vec{b}_{t,\mathcal{O}}^{(j)})^{T}\\ 
&(\vec{A}_{t,\mathcal{O}}^{(j)} \vec{\Sigma}_{t,\mathcal{O}}^{(j)} (\vec{A}_{t,\mathcal{O}}^{(j)})^{T} )^{-1}
(\vec{\xi}_{t,\mathcal{O}}- \vec{A}_{t,\mathcal{O}}^{(j)} \vec{\mu}_{t,\mathcal{O}}^{(j)}-\vec{b}_{t,\mathcal{O}}^{(j)}).
\label{equ:tmp}
\end{aligned}
\end{equation}
By calculating derivatives of (\ref{equ:tmp}) with respect to $\vec{\xi}_{t,\mathcal{O}}$, the optimal solution can be derived, which is equivalent to the product of Gaussians, i.e.,
\begin{equation}
\vec{\xi}_{t,\mathcal{O}} \sim  \prod_{j=1}^{P}
\mathcal{N} \left( \vec{A}_{t,\mathcal{O}}^{(j)} \vec{\mu}_{t,\mathcal{O}}^{(j)}+\vec{b}_{t,\mathcal{O}}^{(j)}, \vec{A}_{t,\mathcal{O}}^{(j)} \vec{\Sigma}_{t,\mathcal{O}}^{(j)} (\vec{A}^{(j)}_{t,\mathcal{O}})^{T} \right),
\label{equ:affine}
\end{equation}
where $\vec{A}^{(j)}_{t,\mathcal{O}}$ and $\vec{b}^{(j)}_{t,\mathcal{O}}$ respectively correspond to the output blocks of $\vec{A}^{(j)}_{t}=blockdiag(\vec{A}^{(j)}_{t,\mathcal{I}} ,\vec{A}^{(j)}_{t,\mathcal{O}})$ and $\vec{b}^{(j)}_{t}=[(\vec{b}^{(j)}_{t,\mathcal{I}})^{T} \;(\vec{b}^{(j)}_{t,\mathcal{O}})^{T}]^{T} $. 
Note that the input blocks $\vec{A}^{(j)}_{t,\mathcal{I}}$ and $\vec{b}^{(j)}_{t,\mathcal{I}}$ are used to retrieve the local desired inputs $\vec{\xi}_{t,\mathcal{I}}^{(j)}$ projected into the different task frames,
which act as the conditional inputs for the generation of local trajectories.
An illustration of trajectory encoding via GMM and trajectory retrieval via GMR is provided in 
Fig.~\ref{fig:gmm:gmr}. It can be observed from Fig.~\ref{fig:gmm:gmr}(\emph{top-right}) that, the first Gaussian component in frame $\{1\}$ and the third component in frame $\{2\}$ (counting from left to right) have the smallest covariances, implying that trajectory segments encapsulated by these components are highly consistent across demonstrations, and therefore represent an important feature of the movements. 
\subsection{Confidence-weighted Task-parameterized Movement Learning \label{subsec:cw:tmp}}
Among previous works on task-parameterized learning \cite{Calinon2014, Leonel, Joao}, task frames and associated parameters $\{\vec{A}_{t}^{(j)},\vec{b}_{t}^{(j)}\}$ were defined beforehand. Moreover, task frames were assigned with equal priorities.
However, it may happen that, for some specific task frames, their influences are expected to be larger than the rest of frames,  
and hence it is desired to introduce human confidence about task frames.
On the basis of (\ref{equ:affine}), the human prior information can be naturally incorporated into task frames.
Assuming that the confidences of different task frames are known, let us denote them as $c_{t,j} \in (0,1)$.  
We then formulate the original objective of the variant of TP-GMM as $\prod_{j=1}^{P} \mathcal{P} \left( \vec{\xi}_{t,\mathcal{O}} | \vec{A}_{t,\mathcal{O}}^{(j)} \vec{\mu}_{t,\mathcal{O}}^{(j)}+\vec{b}_{t,\mathcal{O}}^{(j)}, \vec{A}_{t,\mathcal{O}}^{(j)} \vec{\Sigma}_{t,\mathcal{O}}^{(j)} (\vec{A}_{t,\mathcal{O}}^{(j)})^{T} \right)^{c_{t,j}}$. Here, $c_{t,j}$ can be interpreted as a measurement of the contribution of each local conditional Gaussian distribution to the product operation. 
Similar to the derivation of (\ref{equ:affine}), 
the optimal estimation of $\vec{\xi}_{t,\mathcal{O}}$ can be determined by
\begin{equation}
\vec{\xi}_{t,\mathcal{O}} \!\sim \! \prod_{j=1}^{P}\!
\mathcal{N} \! \left( \vec{A}_{t,\mathcal{O}}^{(j)} \vec{\mu}_{t,\mathcal{O}}^{(j)} \! + \! \vec{b}_{t,\mathcal{O}}^{(j)}, \vec{A}_{t,\mathcal{O}}^{(j)} (\vec{\Sigma}_{t,\mathcal{O}}^{(j)}/c_{t,j}) (\vec{A}_{t,\mathcal{O}}^{(j)})^{T} \right).
\label{equ:conf:affine}
\end{equation}
The above result has an intuitive interpretation: if the frame $\{j\}$ has a higher (lower) confidence $c_{t,j}$ at time $t$, its contribution to the Gaussian product is higher (lower) due to a smaller (larger) covariance , i.e., $\vec{\Sigma}_{t,\mathcal{O}}^{(j)}/c_{t,j}$.
Figure~\ref{fig:gmm:gmr}(\emph{bottom-left}) depicts an example of applying confidence-weighted scheme, where the resulting trajectory favors local trajectory in the task frame that is assigned with a higher confidence.

\section{Optimization of Task-parameterized Movement Trajectories \label{sec:rl_tp}}
In this section we address the question: how can good task parameters be selected?
For instance, for applications with flexible task parameters, i.e., different values of task parameters allow for finishing the same task, which configuration of parameters is better? We tackle this problem by optimizing task parameters from a reinforcement learning perspective (Section~\ref{subsec:rl_tp}), 
and subsequently, we provide a dual perspective on this optimization, so that a connection between our approach and the standard optimization of GMM components is built (Section~\ref{subsec:dual}).

\subsection{Reinforcement Learning of Task Parameters \label{subsec:rl_tp}}
Considering that task parameters $\{\vec{A}_{t}^{(j)}$,$\vec{b}_{t}^{(j)}\}$ describe different task frames (and therefore, different task situations),
a straightforward way to refine them is by applying rotation and translation operations to their pre-defined values. Since the input blocks in task parameters are often uncontrollable (e.g., a time sequence input), we only discuss the learning of output blocks, i.e., $\{\vec{A}_{t,\mathcal{O}}^{(j)}$, $\vec{b}_{t, \mathcal{O}}^{(j)}\}$. 
Formally, let us define new rotation matrices and translational vectors as $\{\vec{R}_{t}^{(j)}\}_{j=1}^{P}$ and   
$\{\vec{d}_{t}^{(j)}\}_{j=1}^{P}$, respectively. 
Then, for an arbitrary local trajectory point $\vec{\xi}_{t,\mathcal{O}}^{({j})}$ in the frame $\{{j}\}$, after new rotational and translational operations are performed, we can prove that its representation in the reference frame $\{O\}$ becomes 
\begin{equation}
\hat{\vec{\xi}}_{t,\mathcal{O}}^{({j})}=\underbrace{{\vec{A}}_{t,\mathcal{O}}^{(j)} \vec{R}_t^{(j)}}_{\widehat{\vec{A}}_{t,\mathcal{O}}^{({j})}} \vec{\xi}_{t,\mathcal{O}}^{({j})} + \underbrace{{\vec{A}}_{t,\mathcal{O}}^{(j)} \vec{d}_{t}^{(j)} +{\vec{b}}_{t,\mathcal{O}}^{(j)}}_{\widehat{\vec{b}}_{t,\mathcal{O}}^{({j})}}.
\label{equ:linear:transform}
\end{equation}
Accordingly, new task parameters $\{\widehat{\vec{A}}_{t,\mathcal{O}}^{({j})},\widehat{\vec{b}}_{t,\mathcal{O}}^{({j})}\}$ of the frame $\{{j}\}$ are determined, which can be later used to replace initial task parameters  $\{\vec{A}_{t,\mathcal{O}}^{(j)}$,$\vec{b}_{t,\mathcal{O}}^{(j)}\}$ and generate a new trajectory sequence in the frame $\{O\}$ via (\ref{equ:affine}). With the affine transformation in (\ref{equ:linear:transform}), we actually learn task parameters by finding the optimal rotation matrices $\vec{R}_{t}^{(j)}$ and translational vectors $\vec{d}_{t}^{(j)}$.
 
We here consider that the rotation matrix represents sequential rotations about $(x,y,z)$ axes with angles $(\alpha,\beta,\gamma)$, and therefore the determination of $\{\vec{R}_t^{(j)}\}_{j=1}^{P}$ is equivalent to that of $\{\alpha_{t}^{(j)},\beta_{t}^{(j)},\gamma_{t}^{(j)}\}_{j=1}^{P}$. 
Furthermore, let us denote $\vec{a}_t^{(j)}=\begin{matrix}[
 \alpha_{t}^{(j)} \!& \! \beta_{t}^{(j)} & \! \gamma_{t}^{(j)} \! & \! \vec{d}_{t}^{(j)T}]
\end{matrix}$ and $\vec{a}_t=\begin{matrix} [\vec{a}_t^{(1)} \! & \! \vec{a}_t^{(2)}\! &\! \cdots&\!\! \vec{a}_t^{(P)}]^{T}\end{matrix}$. In order to formulate the learning of $\vec{a}_t$ into a RL problem, we represent $\vec{a}_t$ as a parametric policy, i.e.,
\begin{equation}
\vec{a}_t=\vec{\Phi}_t(\vec{\theta}+\vec{\epsilon}),
\label{equ:pi_2:parameter}
\end{equation}
where $\vec{\Phi}_t$ and $\vec{\epsilon}$ represent basis functions and stochastic exploration noise, respectively, and $\vec{\theta}$ denotes the policy parameters to be learned.
By optimizing $\vec{\theta}$ with respect to additional constraints (i.e., task-dependent cost functions), the optimal parameters $\vec{a}_t$ can be found, which are subsequently used to retrieve new task parameters based on (\ref{equ:linear:transform}).
Since we focus on optimizing the task parameters associated with task frames, the rotation matrix (defined by rotation angles) is an intuitive way to modify frames.
Also, here we focus on learning positions rather than orientations, and thus this rotation operation suffices for our optimization problem. 

For the typical policy search problem (\ref{equ:pi_2:parameter}), many algorithms have been proven effective.
Here, we take policy improvement with path integrals (PI$^2$) \cite{Buchli,Evangelos} as an example to illustrate the reinforcement learning of task parameters. 
Let us denote the exploration noise at time step $i\in \{1,2,\ldots, N\}$ during the roll-out (i.e., episode) $h\in \{1,2,\ldots,H\}$ as $\vec{\epsilon}_{i,h}$, where $N$ is the time length of a roll-out and $H$ is the number of roll-outs. 
As suggested in \cite{Stulp}, we apply a constant exploration noise $\vec{\epsilon}_{h}$ during the $h$-{th} roll-out (i.e., $\vec{\epsilon}_{h}=\vec{\epsilon}_{i,h}, \forall i\in \{1,2,\ldots, N\}$) and update the policy parameters using every $H$ roll-outs.
In each roll-out, we can first calculate $\vec{a}_t$ using
(\ref{equ:pi_2:parameter}). Subsequently, we can retrieve new task parameters $\{\widehat{\vec{A}}_{t,\mathcal{O}}^{({j})},\widehat{\vec{b}}_{t,\mathcal{O}}^{({j})}\}_{j=1}^{P}$ using (\ref{equ:linear:transform}). By plugging new task parameters and local trajectories in different frames into (\ref{equ:affine}), an updated trajectory in the reference frame can be generated. Moreover, on the basis of the cost function (which is usually pre-defined depending on the specific task and additional constrains), we can compute the cumulative cost value $S_h$ for each roll-out. 
Thus, given the cumulative costs $\{S_h\}_{h=1}^{H}$ in $H$ roll-outs, the policy parameters are updated as follows
\begin{equation}
\vec{\theta}:=\vec{\theta}+\sum_{h=1}^{H} w_h \vec{\epsilon}_{h}
\label{equ:pi_2},
\end{equation}
with
$w_h=\frac{ e^{- \kappa S_h } }{ \sum_{h=1}^{H} e^{- \kappa S_h} }$ and $\kappa>0$.
We can continuously perform explorations and update $\vec{\theta}$ every $H$ roll-outs until $\vec{\theta}$ converges or the cumulative cost is below a certain value. The complete
learning procedure is illustrated in \emph{Algorithm}~\ref{algorithm:optimization}. Figure~ \ref{fig:gmm:gmr}(\emph{bottom-middle}) shows trajectory adaptation using optimized task parameters. As it can be seen, the final trajectory is modulated by using new task parameters.

\subsection{Dual Perspective of Optimizing Task Parameters \label{subsec:dual}}
In this section, we provide a dual perspective to interpret the optimization of task parameters. 
To do so, let us first recall the main result (\ref{equ:reproduce}) in TP-GMM. 
Note that in (\ref{equ:reproduce}), there exists an affine transformation of the GMM component $\{\vec{\mu}_{k}^{(j)},\vec{\Sigma}_{k}^{(j)}\}$ through the task parameters $\{{\vec{A}}_{t}^{(j)}, {\vec{b}}_{t}^{(j)}\}$. 
If we write the block-decomposition of
$\vec{\mu}_{k}^{(j)}=\left[\begin{matrix}
\vec{\mu}_{k,\mathcal{I}}^{(j)} \\ \vec{\mu}_{k,\mathcal{O}}^{(j)}
\end{matrix} \right]$, $\vec{\Sigma}_{k}^{(j)}=\left[\begin{matrix}
\vec{\Sigma}_{k,\mathcal{I}\mathcal{I}}^{(j)} & \vec{\Sigma}_{k,\mathcal{I}\mathcal{O}}^{(j)} \\ 
\vec{\Sigma}_{k,\mathcal{O}\mathcal{I}}^{(j)} & \vec{\Sigma}_{k,\mathcal{O}\mathcal{O}}^{(j)}
\end{matrix} \right]$
and substitute the optimized task parameters $\{\widehat{\vec{A}}_{t,\mathcal{O}}^{(j)}, \widehat{\vec{b}}_{t,\mathcal{O}}^{(j)}\}$ into (\ref{equ:reproduce}), new  
mean and covariance can be derived as
\begin{equation}
\begin{aligned}
&\widehat{\vec{\mu}}_{t,k}^{(j)}=\left[\begin{matrix}
\vec{A}_{t,\mathcal{I}}^{(j)} & \vec{0} \\ 
\vec{0} & \widehat{\vec{A}}_{t,\mathcal{O}}^{(j)}
\end{matrix} \right]
\left[\begin{matrix}
\vec{\mu}_{k,\mathcal{I}}^{(j)} \\ \vec{\mu}_{k,\mathcal{O}}^{(j)}
\end{matrix} \right] 
+ \left[\begin{matrix}
\vec{b}_{t,\mathcal{I}}^{(j)} \\ \widehat{\vec{b}}_{t,\mathcal{O}}^{(j)} 
\end{matrix} \right] \\
&\widehat{\vec{\Sigma}}_{t,k}^{(j)}=\left[\begin{matrix}
\vec{A}_{t,\mathcal{I}}^{(j)} & \vec{0} \\ 
\vec{0} & \widehat{\vec{A}}_{t,\mathcal{O}}^{(j)}
\end{matrix} \right]
\left[\begin{matrix}
\vec{\Sigma}_{k,\mathcal{I}\mathcal{I}}^{(j)} & \vec{\Sigma}_{k,\mathcal{I}\mathcal{O}}^{(j)} \\ 
\vec{\Sigma}_{k,\mathcal{O}\mathcal{I}}^{(j)} & \vec{\Sigma}_{k,\mathcal{O}\mathcal{O}}^{(j)}
\end{matrix} \right]
\left[\begin{matrix}
\vec{A}_{t,\mathcal{I}}^{(j)} & \vec{0} \\ 
\vec{0} & \widehat{\vec{A}}_{t,\mathcal{O}}^{(j)}
\end{matrix} \right]^{T}.
\end{aligned}
\label{equ:dual}
\end{equation} 
This new mean and covariance can also be seen as being equivalent to a new local model $\{\widehat{\vec{\mu}}_{k}^{(j)}, \widehat{\vec{\Sigma}}_{k}^{(j)}\}$, rotated and translated by the old parameters $\{{\vec{A}}_{t}^{(j)}, {\vec{b}}_{t}^{(j)}\}$, resulting in
\begin{equation}
\begin{aligned}
&\widehat{\vec{\mu}}_{t,k}^{(j)}=\left[\begin{matrix}
\vec{A}_{t,\mathcal{I}}^{(j)} & \vec{0} \\ 
\vec{0} & {\vec{A}}_{t,\mathcal{O}}^{(j)}
\end{matrix} \right]             
\left[\begin{matrix}
\widehat{\vec{\mu}}_{k,\mathcal{I}}^{(j)} \\ \widehat{\vec{\mu}}_{k,\mathcal{O}}^{(j)}
\end{matrix} \right] 
+ \left[\begin{matrix}
\vec{b}_{t,\mathcal{I}}^{(j)} \\ {\vec{b}}_{t,\mathcal{O}}^{(j)} 
\end{matrix} \right] \\
&\widehat{\vec{\Sigma}}_{t,k}^{(j)}=\left[\begin{matrix}
\vec{A}_{t,\mathcal{I}}^{(j)} & \vec{0} \\ 
\vec{0} & {\vec{A}}_{t,\mathcal{O}}^{(j)}
\end{matrix} \right]
\left[\begin{matrix}
\widehat{\vec{\Sigma}}_{k,\mathcal{I}\mathcal{I}}^{(j)} & \widehat{\vec{\Sigma}}_{k,\mathcal{I}\mathcal{O}}^{(j)} \\ 
\widehat{\vec{\Sigma}}_{k,\mathcal{O}\mathcal{I}}^{(j)} & \widehat{\vec{\Sigma}}_{k,\mathcal{O}\mathcal{O}}^{(j)}
\end{matrix} \right]
\left[\begin{matrix}
\vec{A}_{t,\mathcal{I}}^{(j)} & \vec{0} \\ 
\vec{0} & {\vec{A}}_{t,\mathcal{O}}^{(j)}
\end{matrix} \right]^{T}.
\end{aligned}
\label{equ:dual:newModel}
\end{equation}
By rewriting both (\ref{equ:dual}) and (\ref{equ:dual:newModel}) in their expanded forms, we have that
\small
\begin{equation}
\begin{aligned}
&\widehat{\vec{\mu}}_{k,\mathcal{I}}^{(j)}
=\vec{\mu}_{k,\mathcal{I}}^{(j)} \\
&\widehat{\vec{\mu}}_{k,\mathcal{O}}^{(j)}
=\left({\vec{A}}_{t,\mathcal{O}}^{(j)}\right)^{-1} \widehat{\vec{A}}_{t,\mathcal{O}}^{(j)} \vec{\mu}_{k,\mathcal{O}}^{(j)} + \left({\vec{A}}_{t,\mathcal{O}}^{(j)}\right)^{-1} \left(\widehat{\vec{b}}_{t,\mathcal{O}}^{(j)} - {\vec{b}}_{t,\mathcal{O}}^{(j)}\right) \\
&\widehat{\vec{\Sigma}}_{k,\mathcal{I}\mathcal{I}}^{(j)}
=\vec{\Sigma}_{k,\mathcal{I}\mathcal{I}}^{(j)} \\
&\widehat{\vec{\Sigma}}_{k,\mathcal{O}\mathcal{I}}^{(j)}
=\left(\vec{A}_{t,\mathcal{O}}^{(j)}\right)^{-1} \widehat{\vec{A}}_{t,\mathcal{O}}^{(j)} \vec{\Sigma}_{k,\mathcal{O}\mathcal{I}}^{(j)} \\
&\widehat{\vec{\Sigma}}_{k,\mathcal{I}\mathcal{O}}^{(j)}
=\left(\widehat{\vec{\Sigma}}_{k,\mathcal{O}\mathcal{I}}^{(j)} \right)^{T}\\
&\widehat{\vec{\Sigma}}_{k,\mathcal{O}\mathcal{O}}^{(j)}
=\left ({\vec{A}}_{t,\mathcal{O}}^{(j)}\right)^{-1} \widehat{\vec{A}}_{t,\mathcal{O}}^{(j)}
\vec{\Sigma}_{k,\mathcal{O}\mathcal{O}}^{(j)} \left( \left({\vec{A}}_{t,\mathcal{O}}^{(j)} \right)^{-1} \widehat{\vec{A}}_{t,\mathcal{O}}^{(j)} \right)^{T}.
\end{aligned}
\label{equ:dual:newValue}
\end{equation}
\normalsize
Thus, the optimization of task parameters (i.e., transform $\{{\vec{A}}_{t,\mathcal{O}}^{(j)}, {\vec{b}}_{t,\mathcal{O}}^{(j)}\}$ into $\{\widehat{\vec{A}}_{t,\mathcal{O}}^{(j)}$, $ \widehat
{\vec{b}}_{t,\mathcal{O}}^{(j)}\}$) is equivalent to the optimization of GMM components (i.e., transform $\{\vec{\mu}_{k}^{(j)},\vec{\Sigma}_{k}^{(j)}\}$ into $\{\widehat{\vec{\mu}}_{k}^{(j)},\widehat{\vec{\Sigma}}_{k}^{(j)}\}$).  
Note that rotation angles and translational vectors in (\ref{equ:linear:transform}) are learned to optimize task parameters. In contrast to classic approaches where GMM parameters (i.e., means, covariances and mixture coefficients) are updated \cite{Guenter}, learning task parameters renders a lower dimensional optimization, which may speed up the learning process. More importantly,
the optimization of task parameters is independent from the model parameters, and thus local consistent features from demonstrations are still maintained, which might be highly desirable for imitation learning.

\begin{algorithm}
	\caption{Optimization of task-parameterized movements}
	\begin{algorithmic}[1]
		\Statex{\textbf{\emph{Initialization}}} 
		\State Define a global reference frame \{O\} and initial candidate task frames $\{j\}_{j=1}^{P}$. 
		\State Collect demonstrations $\{\{\vec{\xi}_{t,m}\}_{t=1}^{N}\}_{m=1}^{M}$. 
	\end{algorithmic}
	\begin{algorithmic}[1]
		\Statex {\emph{\textbf{Phase 1: learn from demonstrations}}}
		\State Project demonstrations into each frame via (\ref{equ:project}) separately.
		\State Fit GMM to projected trajectories in each frame using EM and generate local trajectories from conditional probabilities $\mathcal{P}(\vec{\xi}_{t,\mathcal{O}}^{(j)} | \vec{\xi}_{t,\mathcal{I}}^{(j)})$ in different frames using GMR. 
	\end{algorithmic}
	\begin{algorithmic}[1]
		\Statex {\textbf{\emph{Phase 2: generalization with optimized task parameters}}}
		\State Define new task parameters $\{\{{\vec{A}}_t^{(j)},{\vec{b}}_t^{(j)}\}_{t=1}^{N}\}_{j=1}^{P}$ depending on the new task instance.
		\State Optimize $\{\{{\vec{A}}_t^{(j)},{\vec{b}}_t^{(j)}\}_{t=1}^{N}\}_{j=1}^{P}$ using (\ref{equ:pi_2}) to minimize the cost function defined based on task requirements.
		\State Use optimized task parameters $\{\{\widehat{\vec{A}}_{t,\mathcal{O}}^{(j)},\widehat{\vec{b}}_{t,\mathcal{O}}^{(j)}\}_{t=1}^{N}\}_{j=1}^{P}$, combined with local trajectories in different task frames, to estimate $\vec{\xi}_{t,\mathcal{O}}$
		in $\{O\}$ via (\ref{equ:affine}). 
	\end{algorithmic}
	\label{algorithm:optimization}
\end{algorithm}

\section{Forward Search of Task Frames \label{sec:forward:search}}
Within the task-parameterized learning framework, the number of task frames are usually fixed and pre-determined \cite{Calinon2014, Leonel,Joao}. 
A natural question concerning the number of task frames arises: can we change the number of task frames? More specifically, how can we determine the number of task frames? For instance, in a robot task with many candidate frames, redundant or irrelevant frames might exist, thus it is reasonable to remove these less important task frames so as to alleviate their undesired influences. As shown in Fig.~\ref{fig:gmm:gmr}(\emph{bottom-right}), the task frame $\{2\}$ should be removed since this frame fails to encapsulate any consistent features from demonstrations.
Even though this problem may have a significant impact on the robot performance, it has not been addressed in the previous works.

In analogy to the classical \emph{forward search} used for selecting high dimensional features, we propose an iterative learning scheme to select the most-relevant task frames with respect to additional task constraints (which can be formulated as a cost function). We first consider the trajectory generation using a single frame. Through separate optimization of task parameters of each candidate frame via \emph{Algorithm} 1, we can evaluate the influence of each frame based on their corresponding cost values. Note that the important frames influence the task significantly, and thus their corresponding cost values should decrease rapidly over the learning iterations. In other words the lower the cost, the higher the importance of the frame. With this insight, we can find the best frame in terms of cost values. Subsequently, we consider the trajectory generation using two frames, i.e., the best frame and one from the remaining ones. By evaluating the combination of the best frame and each of the other frames, the optimal two-frames set can be determined. Similarly, we can find the optimal frame set with more frames until the number of task frames reaches the upper limit. 
Note that the frame selection scheme depends on the definition of the cost function, which is closely related to the task requirements.

\begin{figure}[bt] 
\centering 
\includegraphics[width=8cm]{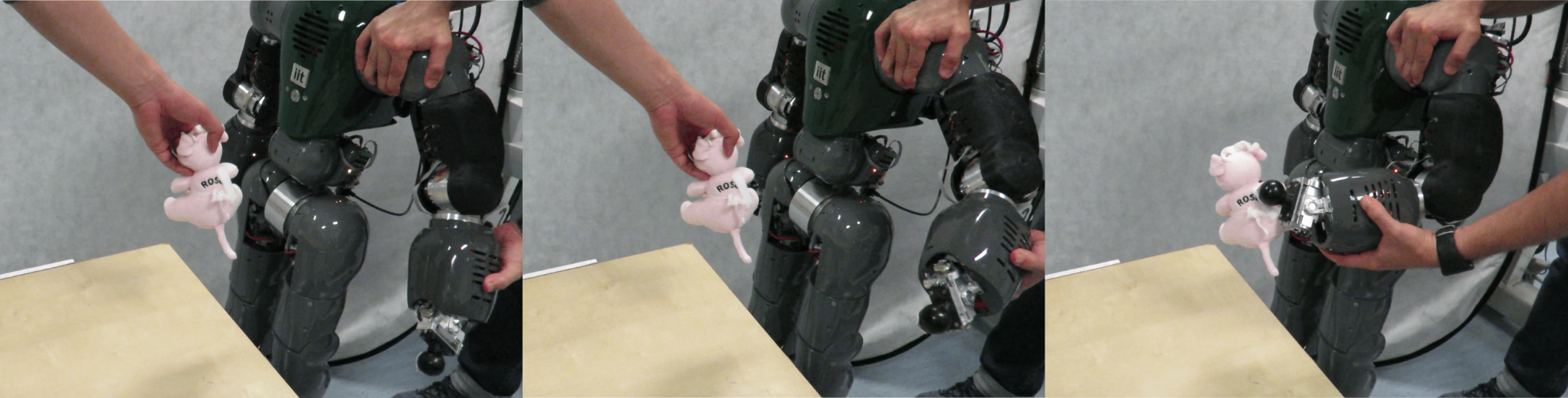} 
\caption{Kinesthetic teaching of a reaching skill on the COMAN robot.}
\label{fig:show_teach_comman} 
\end{figure}

\begin{figure}[bt] \centering 
\includegraphics[width=3.8cm]{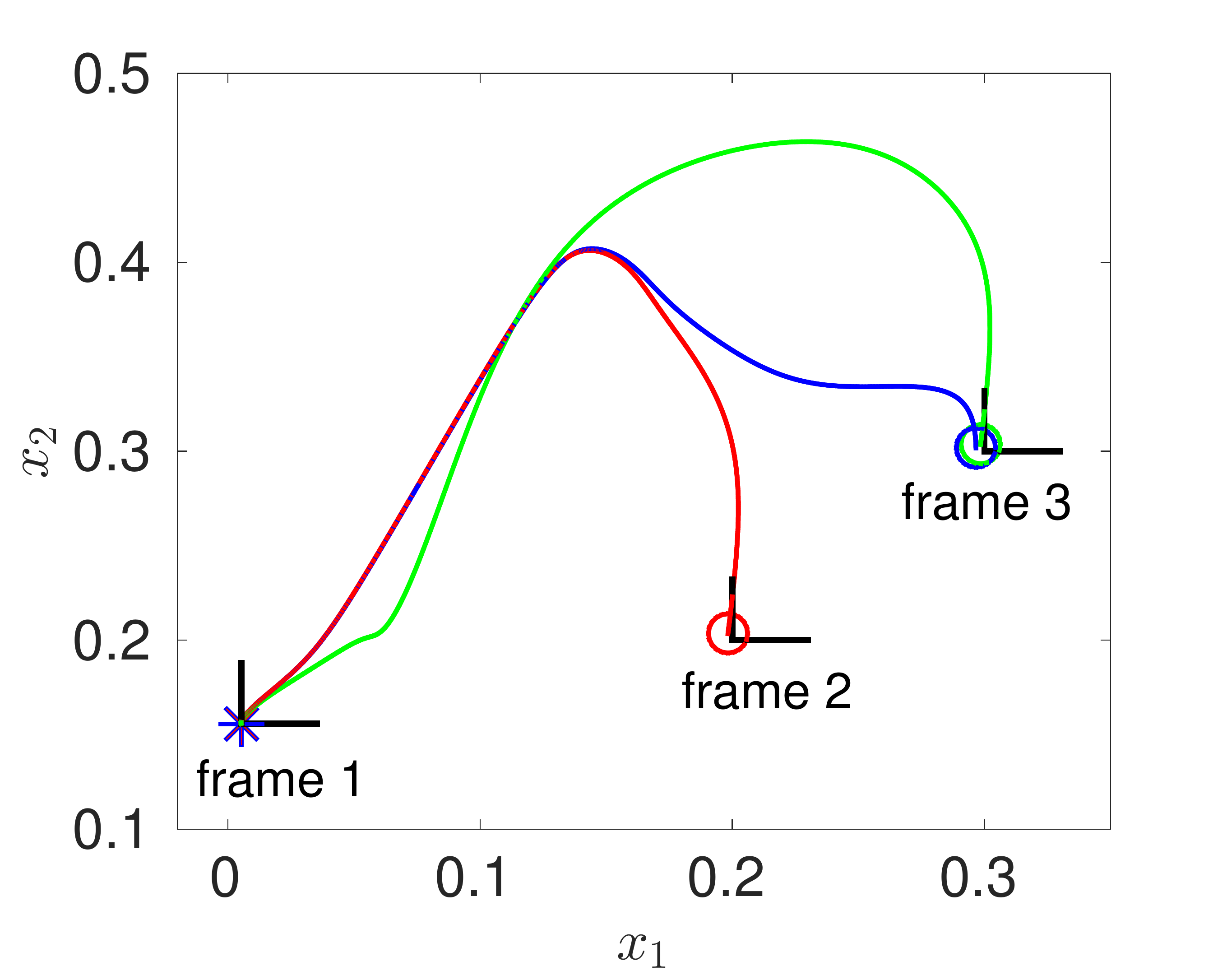}\hspace{0cm}
\includegraphics[width=4.1cm,height=3.05cm]{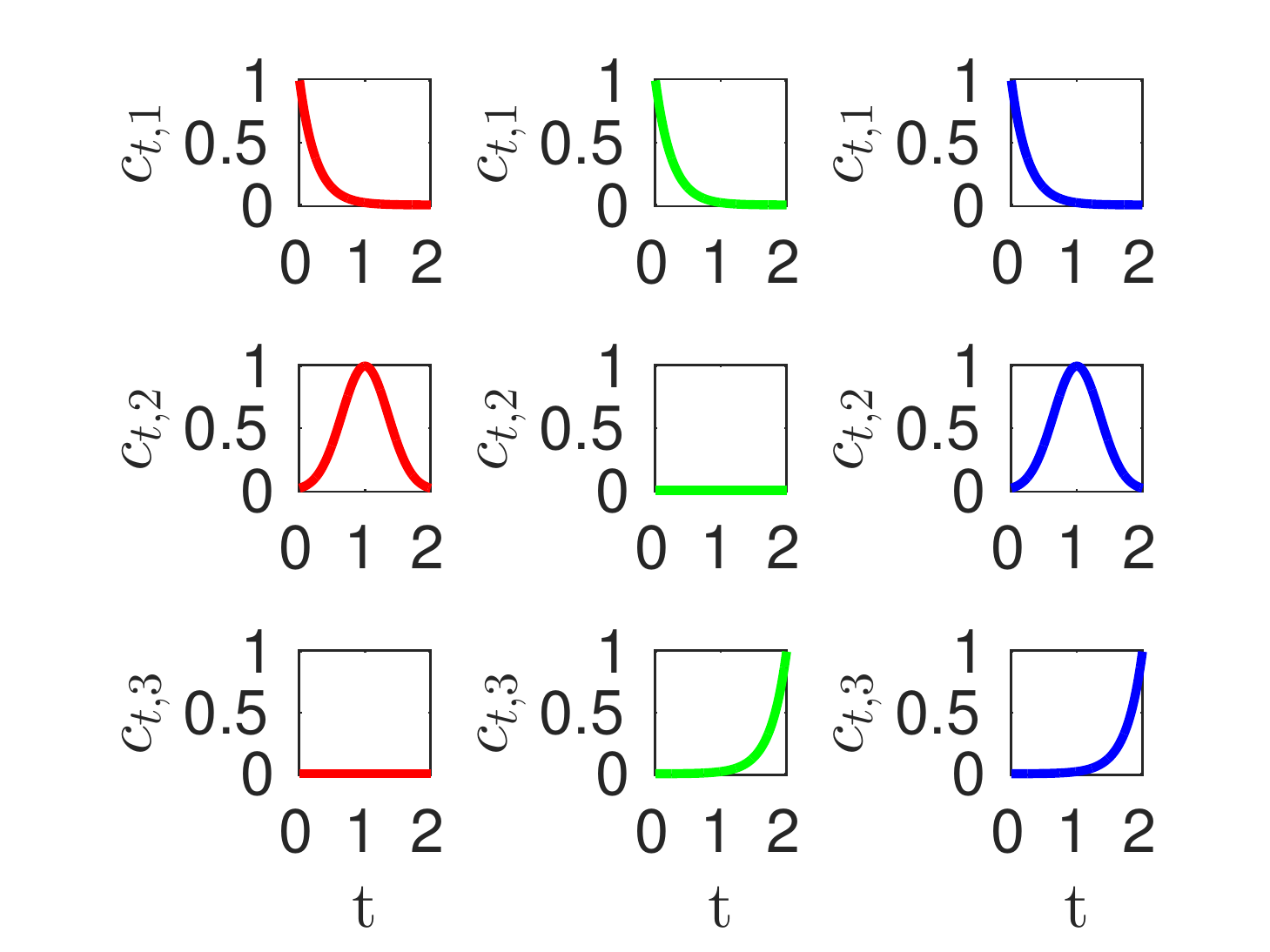}
\caption{Trajectories (\emph{left} graph) generated by assigning different confidences (\emph{right} plot) to task frames. The first, second and third columns of the \emph{right} graph are associated with trajectories depicted by red, green and blue curves in the \emph{left} graph, respectively. 
The start and end point of each trajectory are denoted by `$\ast$' and `$\circ$', respectively.
	}
	\label{fig:TR_TMP} 
\end{figure}

\section{Evaluations \label{sec:evaluations}}
In this section, we evaluate our proposed methods using several examples on the simulated/real COMAN robot \cite{Tsagarakis}: \emph{(i)} we consider the confidence-weighted scheme with different sets of frame confidences (Section~\ref{subsec:evalute:tr-tmp}) in the simulated robot; \emph{(ii)} we apply the task frame optimization to a simulated task of reaching a pole (Section~\ref{subsubsec:pole:task}), a simulated and real tasks of reaching a point (Section~\ref{subsubsec:reach:task}) and a real reaching task with obstacle avoidance (Section~\ref{subsubsec:obs:avoid}); \emph{(iii)} we implement a simulated pick-and-place task to show the frame selection procedure (Section~\ref{subsec:trans:task}). 
Since the tasks are all learned in the robot task space, we use 
the Jacobian matrix $\vec{J}$ of the robot end-effector to control the joint movement, i.e.,
$\widehat{\vec{q}}_{t+1}=\vec{q}_t+\vec{J}(\vec{q}_t)^{\dagger} (\widehat{\vec{p}}_{t+1}-~\vec{p}_{t})$
with $\vec{J}^{\dagger}=\vec{J}^{T}(\vec{J} \vec{J}^T)^{-1}$,
$\vec{q}_t$ and $\vec{p}_t$ respectively represent the joint and Cartesian positions at time $t$, $\widehat{\vec{q}}_{t+1}$ and 
$\widehat{\vec{p}}_{t+1}$ represent the desired positions at time $t+1$.

\subsection{Confidence-weighted Scheme \label{subsec:evalute:tr-tmp}}
We collected 10 reaching trajectories with data-points represented by $\vec{\xi}_t=[\begin{matrix}
t & \vec{p}^{T} \end{matrix}]^{T}$ in the robot base frame $\{O\}$ using kinesthetic teaching on the COMAN's left arm (as shown in Fig.~\ref{fig:show_teach_comman}),
lasting around 2\emph{s} each.
Assuming that the object orientation does not influence the reaching task, we define two initial frames located respectively at the start and end points of each demonstration. Through projecting demonstrations into these frames separately, we train a 4-states GMM to extract local consistency among projected trajectories in each frame, which is after used to retrieve local trajectories using GMR. 
Then, we consider the generalization of local trajectories to new task frames. Note that the robot arm starts from the same position, we therefore use the same start frame, i.e., frame $\{1\}$ in Fig.~\ref{fig:TR_TMP} (\emph{left} graph), described by task parameters ${\vec{A}}^{(1)}=\vec{I}_{4\times4}$ and ${\vec{b}}^{(1)}=[0 \; 0.005\; 0.156\; -0.050]^{T}$. 

In order to illustrate the impact of frame confidences, we consider two new targets which are respectively located at $[0.2\; 0.2\; 0.2]^{T}$ and $[0.3\; 0.3\; 0.2]^{T}$, and thus we define two corresponding target frames $\{2\}$ and $\{3\}$ represented by task parameters ${\vec{A}}^{(2)}={\vec{A}}^{(3)}=\vec{I}_{4\times4}$, ${\vec{b}}^{(2)}=[0 \;0.2\; 0.2 \;0.2]^{T}$, ${\vec{b}}^{(3)}=[0 \;0.3\; 0.3 \;0.2]^{T}$. Three different groups of frame confidences are evaluated, where each group corresponds to a column in Fig.~\ref{fig:TR_TMP} (\emph{right} graph). The final trajectories in $\{O\}$ are computed using (\ref{equ:conf:affine}) and illustrated in the Fig.~\ref{fig:TR_TMP} (\emph{left} graph). Observe that the blue curve coincides with the red one at the beginning and gradually moves towards the green one, implying that   
frame confidences determine the contributions of frames, i.e., the frame assigned with a large (small) confidence has a large (small) influence on the final trajectory. 
It is worth pointing out that the confidence-weighted scheme provides a straightforward way to incorporate additional human experience (if available) about the importance of task frames, whereas the original formulation of TP-GMM does not address this functionality.

\subsection{Task Frame Optimization\label{subsec:opti:task:para}}
Here we show the experiments corresponding to the optimization of task parameters (\emph{Algorithm} 1), where the same set of demonstrations of the reaching skill introduced in Section~\ref{subsec:evalute:tr-tmp} are used. The optimization is carried out under three different scenarios: (\emph{i}) reaching a pole (Section~\ref{subsubsec:pole:task}), (\emph{ii}) reaching a point (Section~\ref{subsubsec:reach:task}),
and (\emph{iii}) obstacle avoidance (Section~\ref{subsubsec:obs:avoid}). In task (\emph{i}), we compare our approach with TP-GMM \cite{Calinon2014} to show the importance of optimizing  task frames. In task (\emph{ii}), we compare our work with optimization of GMM \cite{Guenter} to show the efficiency of our method. In task (\emph{iii}) we show that obstacle avoidance can be achieved by optimizing task parameters.

\begin{figure}[bt] \centering
	\includegraphics[width=5cm,bb=0cm 0.2cm 21cm 17cm,clip]{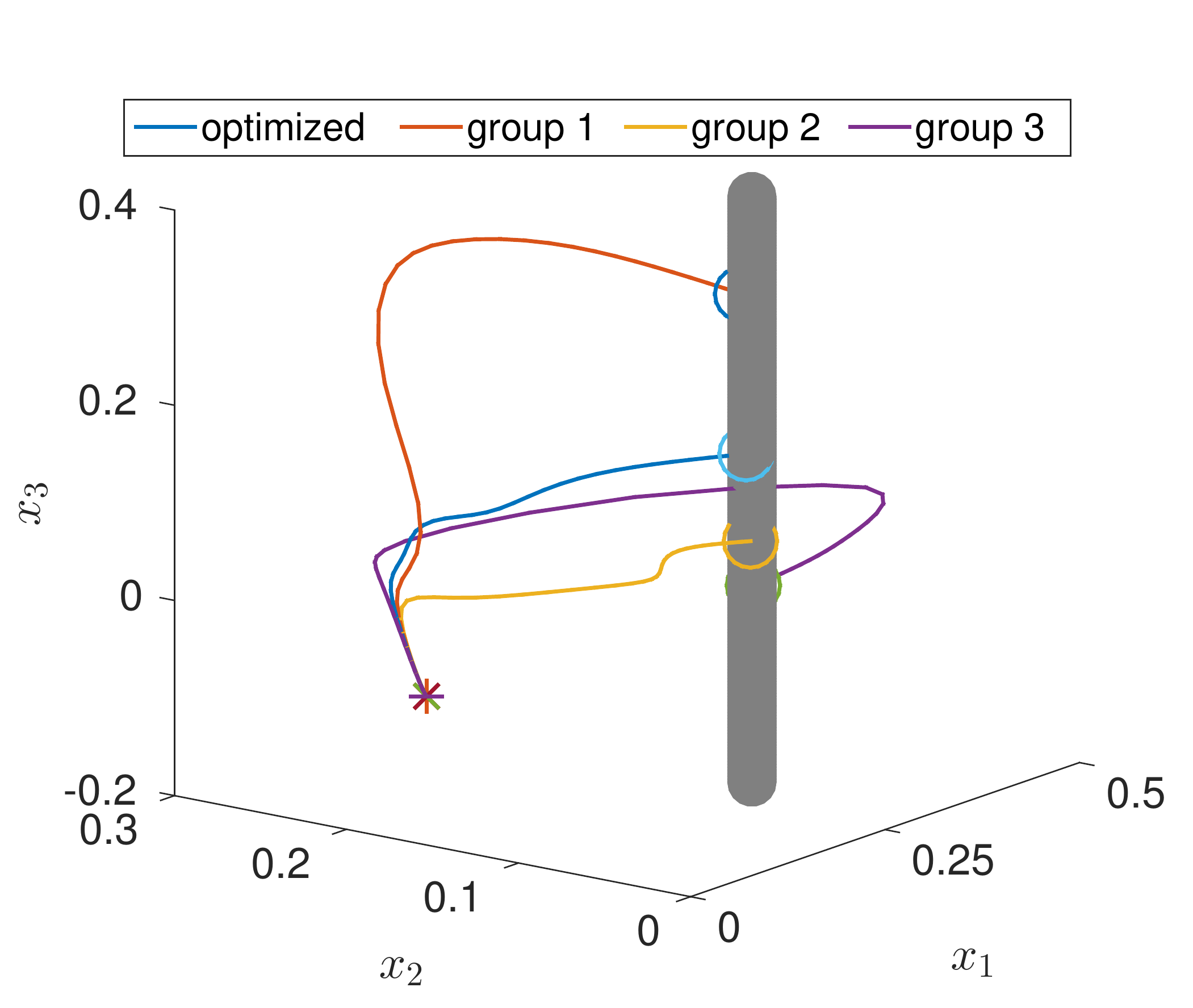}
	\caption{Trajectories generated by using different task parameters from Table~\ref{fig:table} in the task of reaching a static pole.} 
	\label{fig:pole_task} 
\end{figure}

\begin{table*}[bt]
	\caption {Task Parameters of Frame $\{2\}$}
	\vspace{-0.3cm}
	\centering
	\scalebox{1}{	
		\begin{tabular}{ clccc}
			\toprule %
			& \quad \;Group 1 & Group 2 & Group 3 & Optimized Parameters\\ \toprule %
			$\vec{A}^{(2)}$ \!\!
			& \quad \;\;$\vec{I}_{4 \times 4}$ \!
			& $\left[ \begin{matrix}
			1 & 0& 0&0\\0& 0 &-1 &0 \\0& 1 &0& 0 \\0&0&0&1
			\end{matrix}\right]$ \!
			& $\left[ \begin{matrix}
			1 & 0& 0&0\\0&0 &1 &0 \\0& -1 &0& 0 \\0& 0&0&1
			\end{matrix}\right]$\!
			& $\left[ \begin{matrix}
			1 & 0& 0&0\\ 0& 0.5477  & -0.6153   &-0.5670\\ 0& 0.7500  &  0.6615   & 0.0067\\0& 0.3709 &  -0.4288  &  0.8237
			\end{matrix}\right]$\!
			\\
			\midrule
			$\vec{b}^{(2)}$\!\!
			& $[0\;0.3\;0.1\;0.3]^{T}$ \!
			& $[0\;0.3\;0.1\;0.05]^{T}$ \!
			& $[0\;0.3\;0.1\;0]^{T}$\!
			& $[0\;0.3\;0.1\;0.14]^{T}$\!
			\\
			\bottomrule
		\end{tabular}}
		\label{fig:table}
	\end{table*}

\begin{figure}[bt] \centering
	\includegraphics[width=4cm,height=3.3cm]{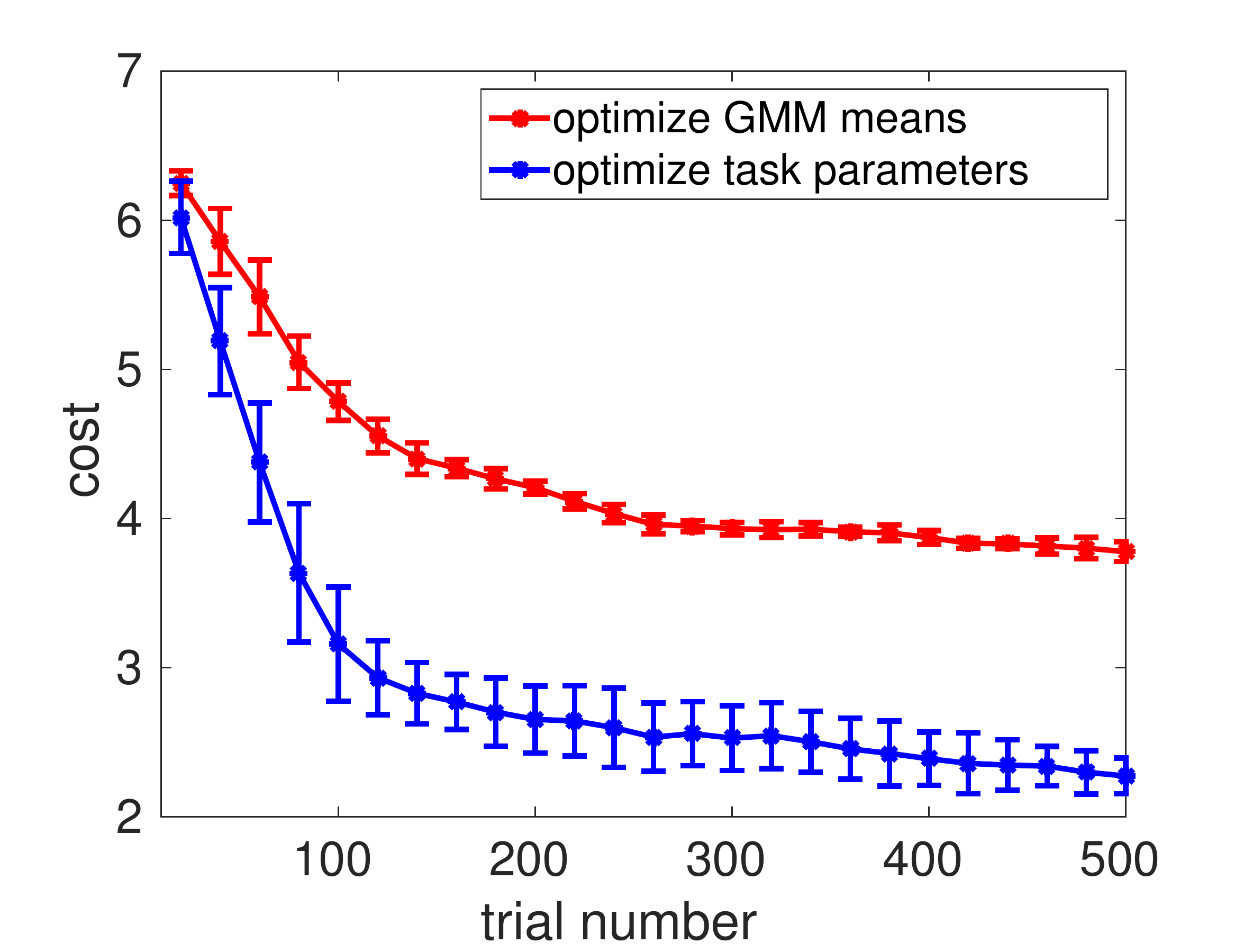}
	\includegraphics[width=4cm,height=3.3cm]{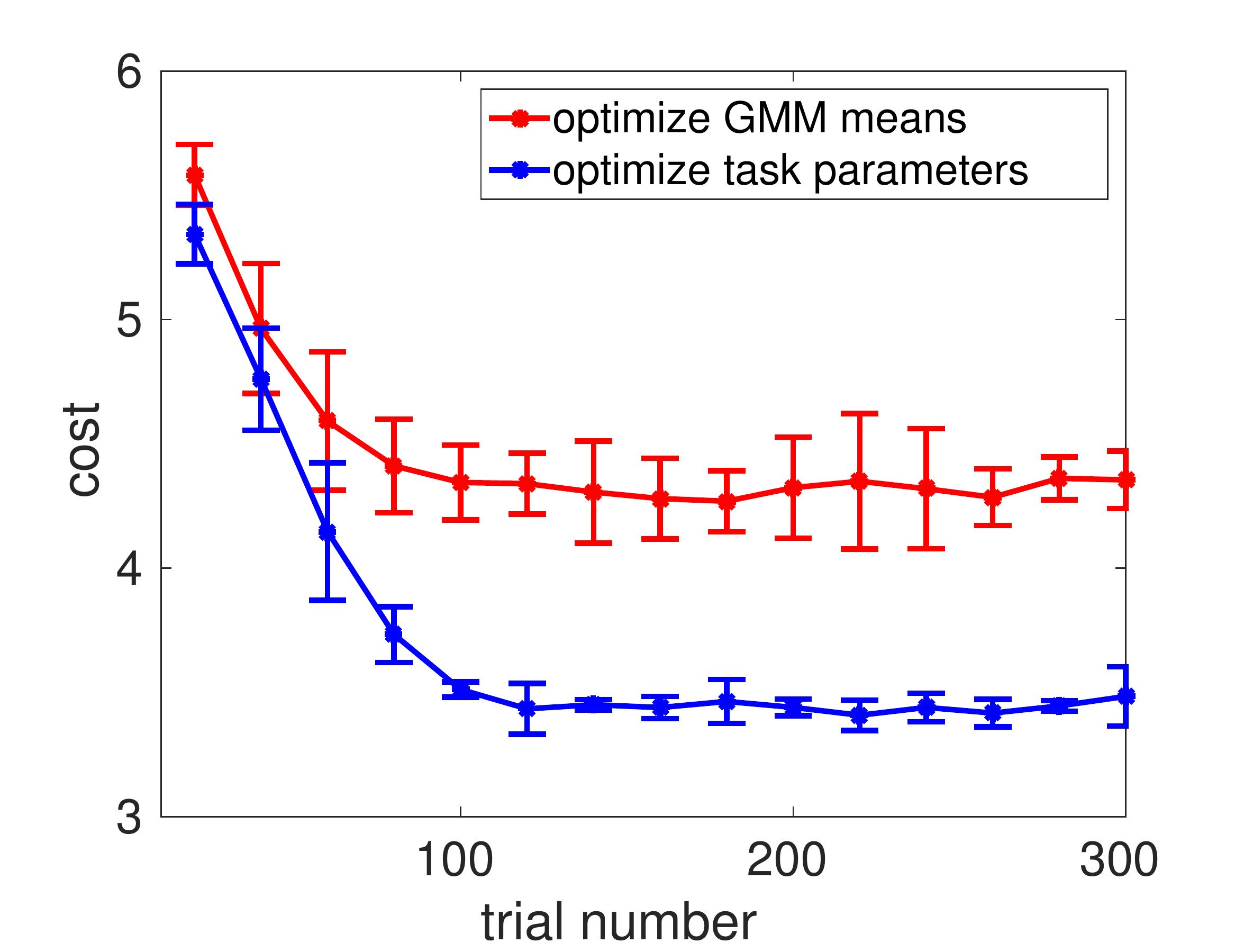}
	\caption{These graphs show 
		cost values of optimizing GMM means and task parameters in different reaching tasks, where the \emph{left} and \emph{right} figures correspond to the target object located at $
		[0.3\;0.3 \; 0.2]^{T}$ and $[-0.3 \; 0.2 \; 0.2]^{T}$ in frame $\{O\}$, respectively. Error-bars represent means and standard deviations of cost values.} 
	\label{fig:pointTask_factor_error} 
\end{figure}

\begin{figure*}[bt] \centering 
\mbox{\includegraphics[height=7cm]{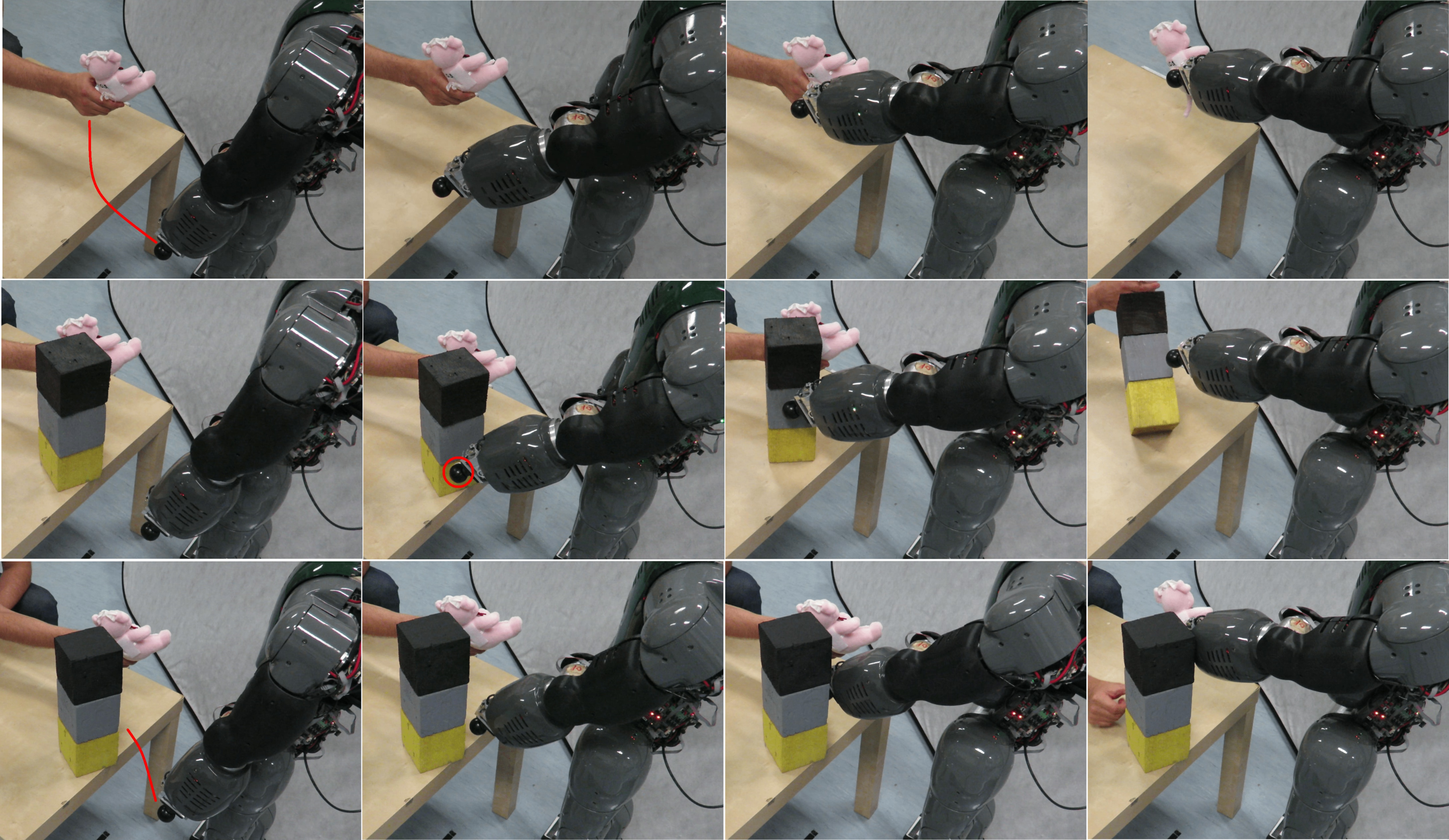}}
	\caption{
		\textit{Top row}: reaching task learned by optimizing task parameters with respect to the cost function (\ref{equ:cost:joint}). \textit{Middle row}: reaching task with obstacle collision, where only the constraint (\ref{equ:cost:joint}) is used. \textit{Bottom row}: reaching task learned by optimizing task parameters with respect to the cost function (\ref{equ:avoid:obs}).
	}
	\label{fig:real:robot} 
\end{figure*}

\subsubsection{Reaching a Pole \label{subsubsec:pole:task}}
We consider the reaching of a static pole that is located at $[0.3\;0.1]$ with the height ranging from $-0.2$ to $0.4$, and the robot only needs to reach it regardless of the exact location along the vertical axis.
Here, we introduce an additional constraint in the joint space, which minimizes the weighted joint displacement, i.e., 
\begin{equation}
f_q= \sum_{t=1}^{N-1} || \vec{W} (\vec{q}_{t+1}-\vec{q}_{t}) ||
\label{equ:cost:joint}
\end{equation}
where $\vec{W}$
denotes a weight matrix and $||\cdot||$ represents 2-norm.
In order to illustrate the importance of optimizing task parameters, we take TP-GMM as a comparison. Since the robot starts from the same position, we define frame $\{1\}$ with parameters $\vec{A}^{(1)}=\vec{I}_{4 \times 4}$, ${\vec{b}}^{(1)}=[0 \; 0.005\; 0.156\; -0.050]^{T}$. Note that the reaching point can be located at any position along the pole vertical axis, so we define three groups of task parameters for frame $\{2\}$, as shown in Table~\ref{fig:table}, which are compared with the optimized task parameters later.
In contrast to these manually defined parameters, we employ our approach (introduced in Section~\ref{sec:rl_tp}) to search optimal task parameters (i.e., the rotation operation and the vertical component in the translational operation) for frame $\{2\}$, where 500 trials are carried out and the policy parameters in (\ref{equ:pi_2}) are updated every 10 trials. The optimal task parameters, found by our approach, are presented in Table~\ref{fig:table}.
The Cartesian trajectories that are generated by using task parameters from Table~\ref{fig:table} 
are depicted in Fig.~\ref{fig:pole_task}. The cost values of using task parameters from group 1, group 2 and group 3 are  
5.21, 5.87 and 5.70, respectively. The optimal task parameters have the cost value 3.26. As it can be seen, it is difficult to manually define appropriate task parameters in TP-GMM. Instead, our method provides an effective way to set task parameters while addressing additional requirements.

\subsubsection{Reaching a Point \label{subsubsec:reach:task}}
In this task, we consider the same joint constraint (\ref{equ:cost:joint}). 
Now, let us first consider the reaching of an object at $
[0.3\;0.3 \; 0.2]^{T}$ in frame $\{O\}$. We define two task frames described by ${\vec{A}}^{(1)}={\vec{A}}^{(2)}=\vec{I}_{4\times4}$, ${\vec{b}}^{(1)}=[0 \; 0.005\; 0.156\; -0.050]^{T}$, ${\vec{b}}^{(2)}=[0 \;0.3\; 0.3 \;0.2]^{T}$.
Note that both frames have their origins at the start and target points respectively, thus only the rotation operations are implemented. 
We use our approach to learn rotation angles for both frames simultaneously, where the policy parameters are updated every 10 roll-outs. For comparison purposes, we also evaluate the optimization of GMM components using PI$^2$. Here, similar to \cite{Guenter}, we optimize Gaussian means while the higher dimensional covariance matrices are kept fixed. 
Besides this current target, we evaluate both optimizations again using a different target located at $[-0.3 \; 0.2 \; 0.2]^{T}$ in $\{O\}$. We have 5 runs for each method and for each target. Meanwhile, we calculate
the average cost every 20 roll-outs in each run. Finally, the means and standard deviations of average costs are computed, as shown by the error-bar curves in Fig.~\ref{fig:pointTask_factor_error}.
Our approach converges faster than GMM components optimization. This result coincides with our intuitions since the frame-based optimization has fewer parameters compared with the GMM optimization. In addition to these evaluations, we test the reaching task on the real COMAN robot, as shown in Fig.~\ref{fig:real:robot} (\emph{top} row), showing that the proposed algorithm generates a trajectory that allows COMAN to perform successfully.

\subsubsection{Obstacle Avoidance\label{subsubsec:obs:avoid}}
We here consider the case in which an obstacle occupies a portion of the learned robot movement path (shown in Fig.~\ref{fig:real:robot}). 
In order to formulate the cost function easily, we simplify the obstacle as a bounded rectangle $S$ and the robot end-effector as a point. Subsequently, we estimate the intersection point $\tilde{\vec{p}}$ of the end-effector trajectory and $S$, and determine if the intersection point lies inside or outside $S$. Furthermore, let us denote the distance between $\tilde{\vec{p}}$ and each edge of $S$ as $d_i$ with $i\in\{1,2,3,4\}$. Then, the cost function can be defined as 
\begin{equation}
C=\left\{
\begin{array}{rl}
f_{q}+k_1e^{(k_{2} d)}, \quad \tilde{\vec{p}} \in S\\ 
f_{q}+k_3e^{(-k_{4} d)}, \;\, \tilde{\vec{p}}  \notin S
\end{array},
\right.
\label{equ:avoid:obs}
\end{equation}
\normalsize
where $d=\min\{d_1,d_2,d_3,d_4\}$ and $k_{i}>0$.
Here, the joint constraint is used to avoid large trajectory deviation from the original desired trajectory.
As a comparison, we test the reaching task using only the cost function (\ref{equ:cost:joint}). The evaluations on the real robot are illustrated in Fig.~\ref{fig:real:robot} (\emph{middle} and \emph{bottom} rows), showing that the robot is capable of both avoiding the obstacle and reaching the target object by optimizing task parameters with respect to (\ref{equ:avoid:obs}).

\begin{figure}[bt] \centering 
	\includegraphics[width=4.1cm,height=3.3cm]{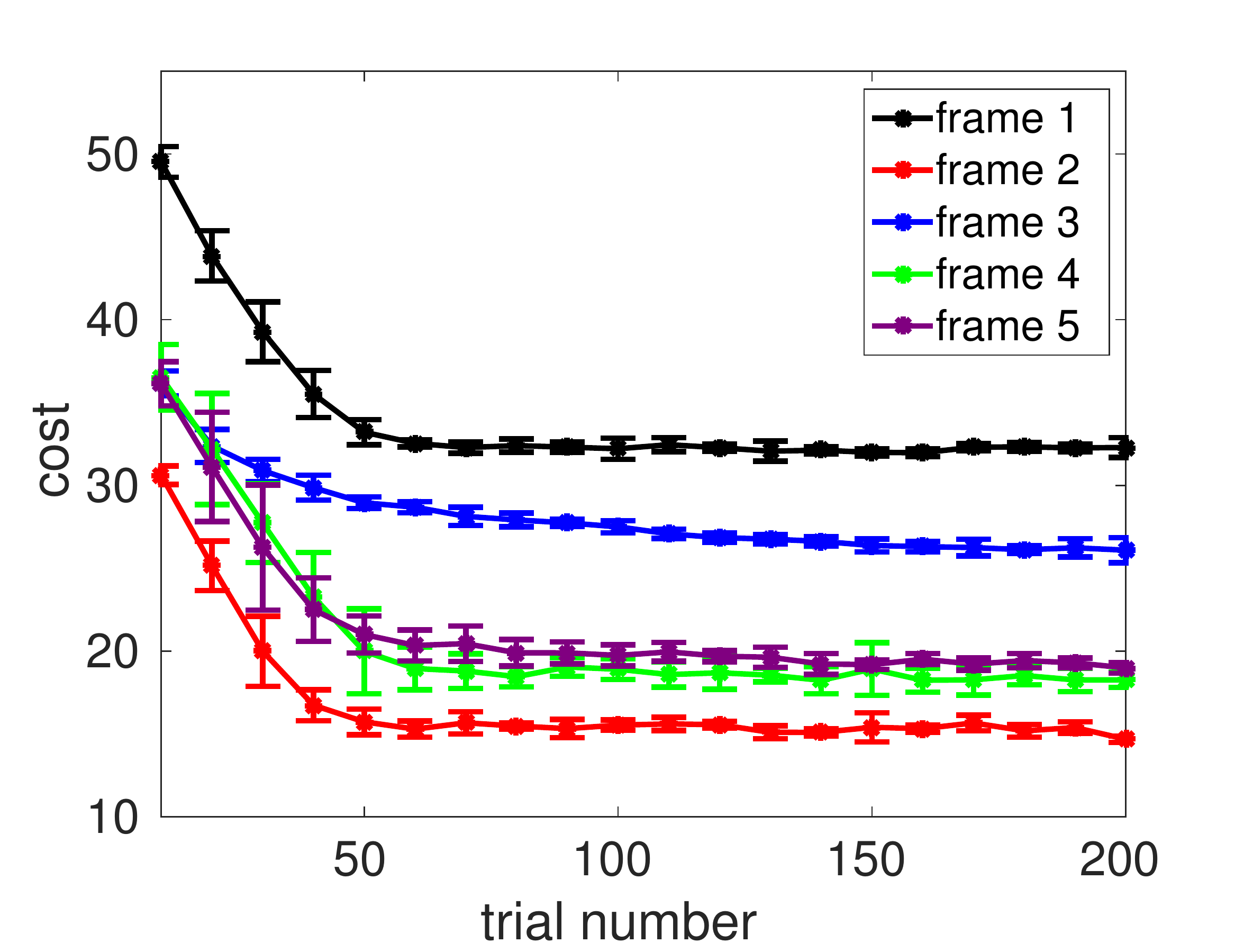} \hspace{0.0cm}
	\includegraphics[width=4.1cm,height=3.3cm]{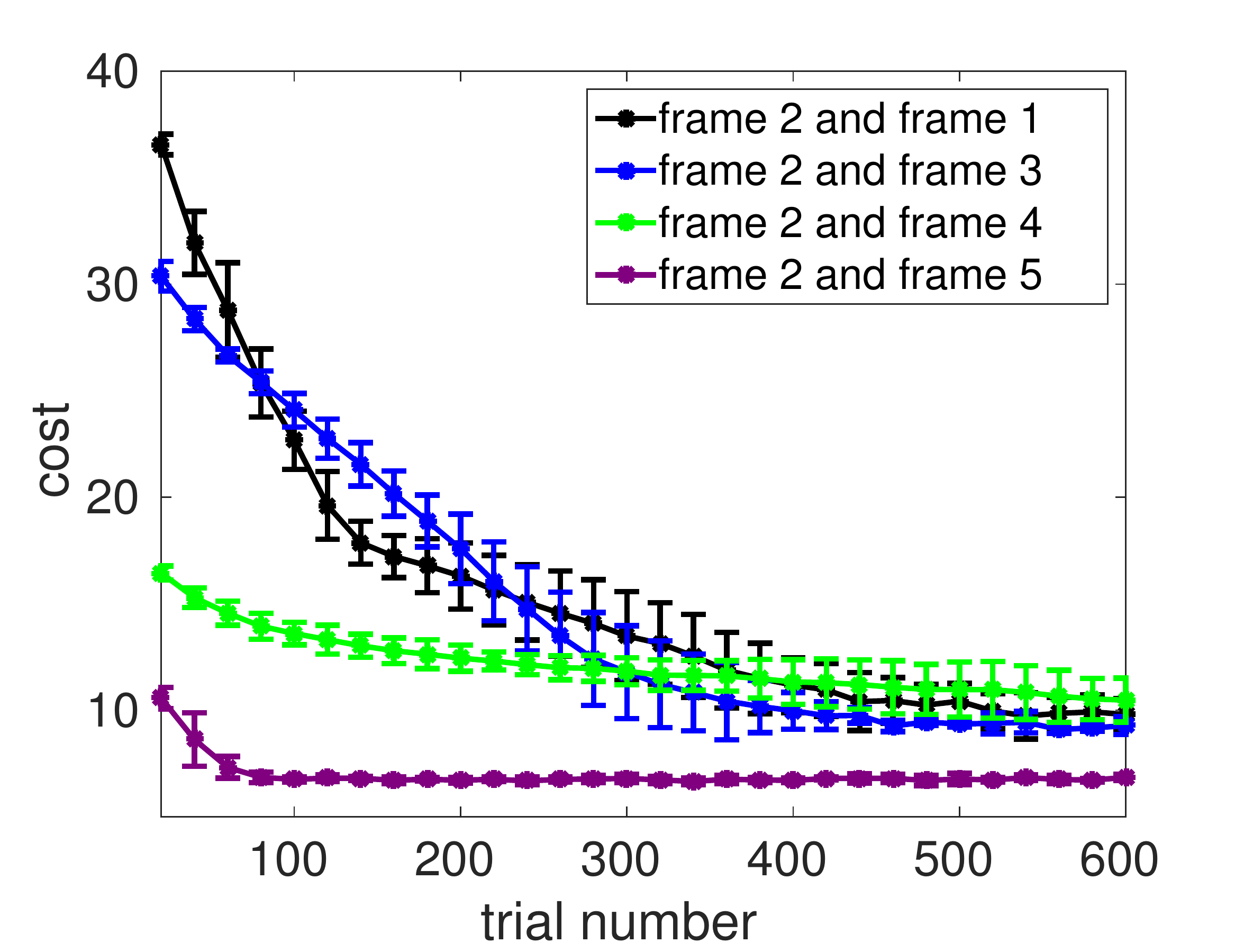} 
	\caption{These figures depict cost values in the frame selection. 
		The \emph{left} plot shows cost values through optimizing a single task frame while the \emph{right} plot shows cost values of optimizing the best frame $\{2\}$ and each of the rest frames. Error-bars represents means and standard deviations of cost values.}
	\label{fig:sort:result} 
\end{figure}

\begin{figure}[bt] \centering 
	\includegraphics[width=4.1cm,height=3.3cm]{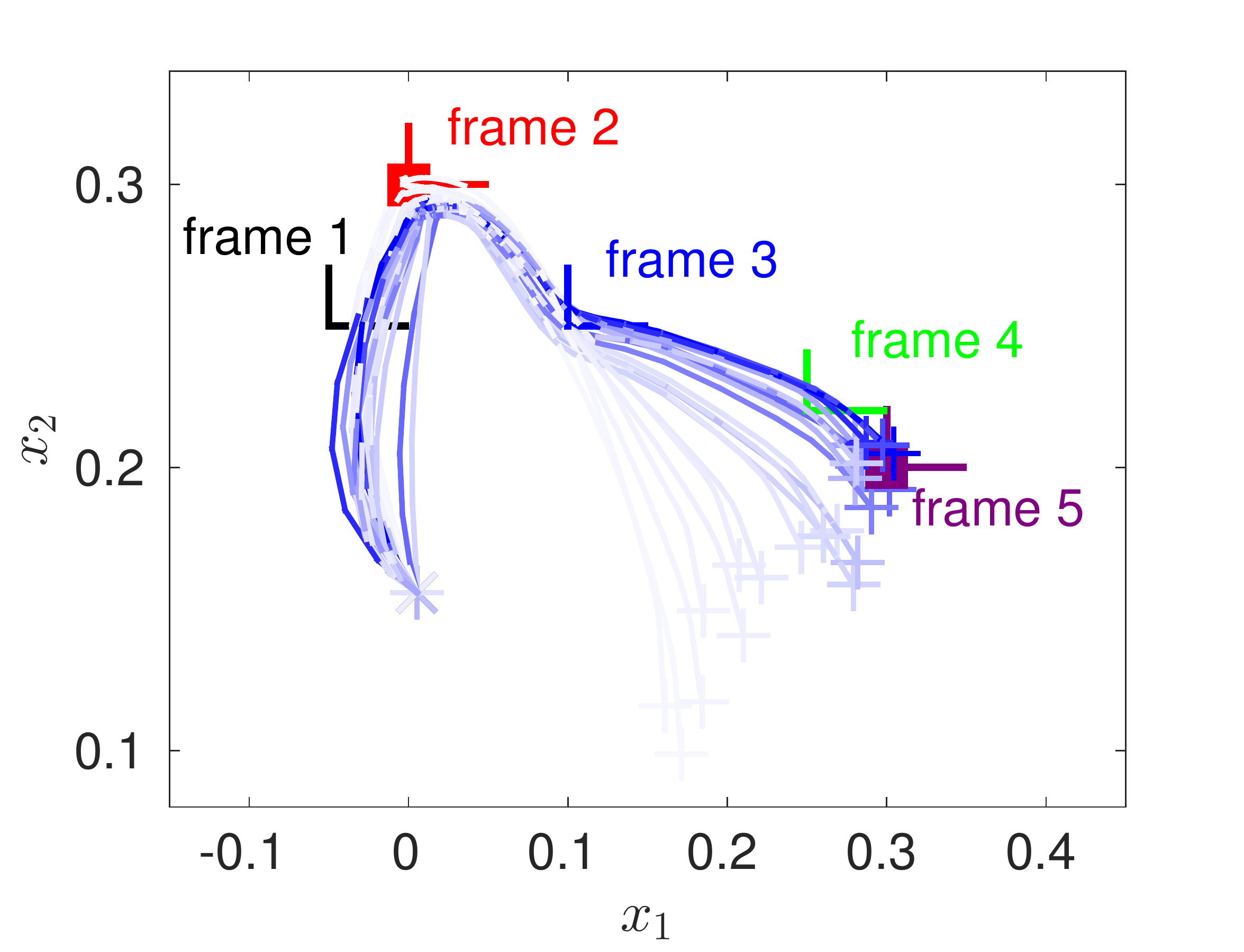}
	\includegraphics[width=4.1cm,height=3.3cm]{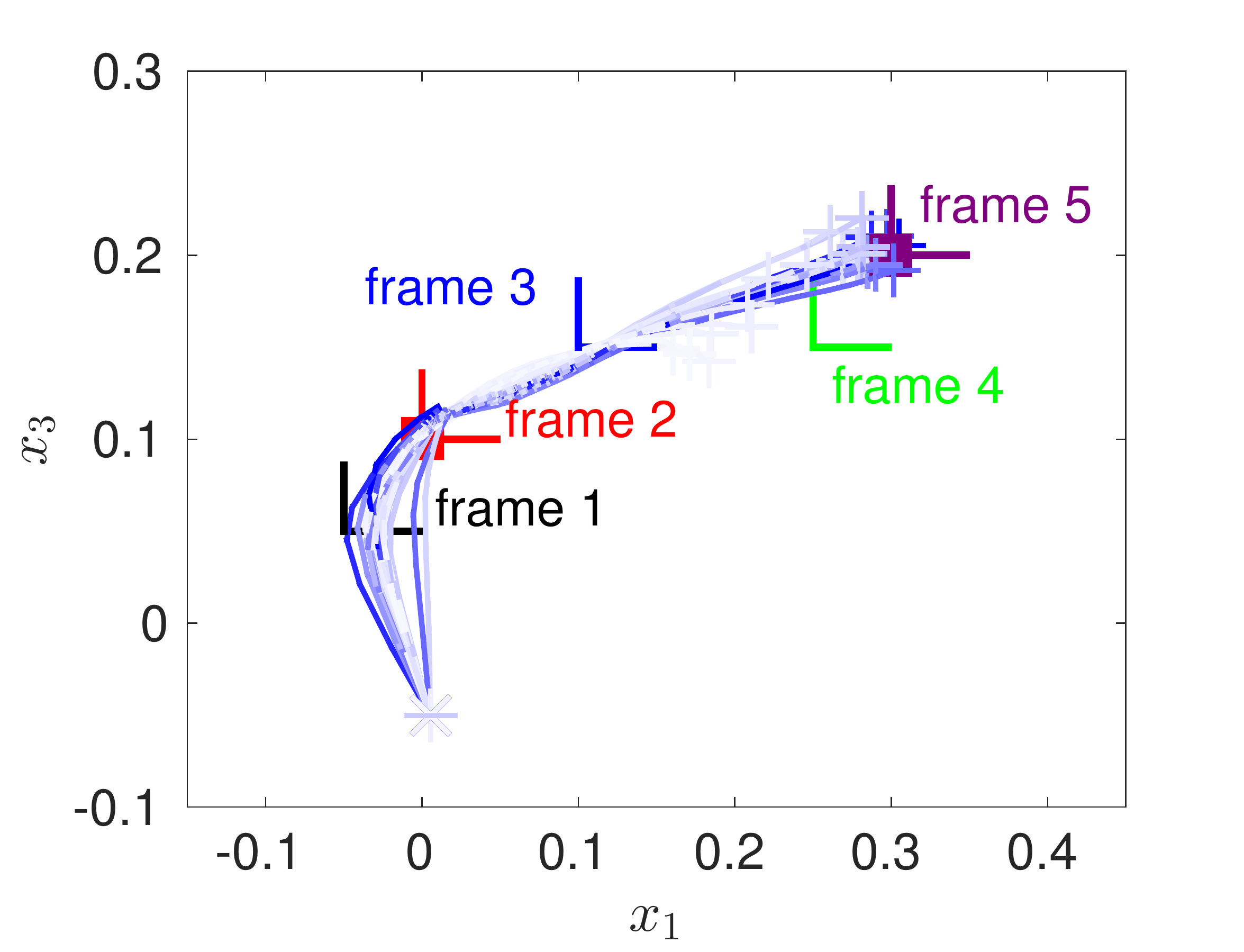} 
	\caption{These figures show the trajectory evolutions (color: from light to dark) through optimizing task parameters of frame $\{2\}$ and frame $\{3\}$ simultaneously. The start and end point of each trajectory are depicted by the blue `$\ast$' and `+', respectively. The red and purple solid boxes denote the desired via-point and end-point.}
	\label{fig:sort:traj} 
\end{figure}

\subsection{Automatic Frame Selection\label{subsec:trans:task}}
We have evaluated a fixed set of task frames in the experiments previously reported, now we consider a transportation task to show the application of the proposed frame selection scheme given a large set of candidate frames. We collected 8 demonstrations of the task through kinesthetic teaching, which guided the robot to reach and pick up an object, and subsequently release it at the goal position, lasting about 10$s$ each. 
We defined 5 initial candidate frames associated with the end-effector positions at time steps
2$s$, 4$s$, 7$s$, 9.5$s$ and 10$s$, respectively. Then, we projected the demonstrations into these candidate frames, and subsequently trained local GMMs and generated local trajectories. 

Considering a new task instance, which requires the robot to pick up the object located at $\vec{p}_s=[0 \;0.3 \; 0.1]^{T}$ (in frame $\{O\}$) when $t=4s$, and subsequently release it at $\vec{p}_e=
[0.3 \;0.2 \; 0.2]^{T}$ (in frame $\{O\}$) when $t=10s$. 
Meanwhile, we expect to reduce the joint displacement. 
Through combining the task and joint constrains, the cost function is defined as 
\begin{equation}
C=f_q+k_{p1}|| \vec{p}_{t=4}-\vec{p}_s|| + k_{p2}  ||\vec{p}_{t=10}-\vec{p}_e||,
\label{equ:cost:via_point}
\end{equation}
where 
$k_{p1}$ and $k_{p2}$ are positive scalars. 
For this new task, we adapt 5 candidate frames using new task parameters
${\vec{A}}^{(j)}=\vec{I}_{4\times4}$, $j=\{1,2,\ldots,5\}$, 
${\vec{b}}^{(1)}=[ 0 \; -0.05 \;0.25 \;0.05  ]^{T}$, 
${\vec{b}}^{(2)}=[ 0\;  0.0  \;0.3  \;0.1  ]^{T}$,
${\vec{b}}^{(3)}=[ 0\;  0.10 \;0.25 \;0.15  ]^{T}$, 
${\vec{b}}^{(4)}=[ 0\;  0.25 \;0.22 \;0.15  ]^{T}$ and 
${\vec{b}}^{(5)}=[ 0\;  0.3  \;0.2  \;0.2  ]^{T}$. 
Since these candidate frames only differ in their origins, it is not needed to apply translational operations, 
and hence we focus on rotation operations. 
We first evaluate each frame separately as shown in Fig.~\ref{fig:sort:result} (\emph{left} plot). Since the frame $\{2\}$ has the most significant influence on the pick-and-place task (i.e., the smallest cost values and the fastest convergence speed), 
it is viewed as the most important frame.
Furthermore, we evaluate the combination of frame $\{2\}$ and the rest of frames. An illustration of trajectory evolutions in one run through combined optimization of frame $\{2\}$ and $\{3\}$ is reported in Fig.~\ref{fig:sort:traj}. The evaluations of two frames are shown in Fig.~\ref{fig:sort:result} (\emph{right} graph), showing that 
the combined performance of frame $\{2\}$ and frame $\{5\}$ attains the lowest cost values.
In summary, 
the frame forward search provides an optimal solution to define a frame set achieving lowest cost values. Finally, we emphasize that humans usually have better understanding of task goals than task frames, and thus the strategy of frame selection offers an alternative solution to discover the most task-relevant frames automatically.

\section{Conclusions \label{sec:conclusion}}

In this paper we presented a generalized task-parameterized learning framework, which is initially learned from human demonstrations. 
The generalization first considers the confidence-weighted scheme, which allows for the incorporation of human prior knowledge on the task frames into a variant of TP-GMM. Subsequently, a novel learning perspective is proposed, which directly optimizes task parameters instead of GMM components, rendering a lower dimensional optimization problem. 
Moreover, an iterative feature selection scheme is proposed, which has shown effective to select important task frames and remove frames that are either redundant or irrelevant for the task. 
In our evaluations, we learn task parameters of different frames without considering their correlations.
However, in many tasks (e.g., the robot bimanual task) the task frames are often relevant to each other, and thus the correlations between frames could be exploited, which might help to accelerate the learning process. 
In addition, since various movement primitives such as non-parametric \cite{Huang2017} and parametric \cite{Paraschos} formulations have been developed, a comprehensive comparison needs further exploitation.

\end{document}